\documentclass[10pt,twocolumn,letterpaper]{article}

\usepackage{iccv}
\usepackage{times}
\usepackage{epsfig}
\usepackage{graphicx}
\usepackage{amsmath}
\usepackage{amssymb}
\usepackage{algorithmic}
\usepackage{algorithm}
\DeclareMathOperator*{\argmax}{arg\,max}

\usepackage[pagebackref=true,breaklinks=true,letterpaper=true,colorlinks,bookmarks=false]{hyperref}

\iccvfinalcopy 

\def\iccvPaperID{6} 

\ificcvfinal\fi

\begin{document}

\title{Semi-Automatic Annotation For Visual Object Tracking}

\author{Kutalmis Gokalp Ince$^{(a)}$, Aybora Koksal$^{(a,b)}$, Arda Fazla$^{(b)}$,  A. Aydin Alatan$^{(a,b)}$\\
Center for Image Analysis (OGAM)$^{(a)}$, Department of Electrical and Electronics Engineering$^{(b)}$\\
Middle East Technical University, Ankara, Turkey\\
{\tt\small kutalmis, aybora, arda.fazla, alatan @ metu.edu.tr}
}

\maketitle
\ificcvfinal\thispagestyle{empty}\fi

\begin{abstract}
   We propose a semi-automatic bounding box annotation method for visual object tracking by utilizing temporal information with a tracking-by-detection approach. For detection, we use an off-the-shelf object detector which is trained iteratively with the annotations generated by the proposed method, and we perform object detection on each frame independently. We employ Multiple Hypothesis Tracking (MHT) to exploit temporal information and to reduce the number of false-positives which makes it possible to use lower objectness thresholds for detection to increase recall. The tracklets formed by MHT are evaluated by human operators to enlarge the training set. This novel incremental learning approach helps to perform annotation iteratively. The experiments performed on AUTH Multidrone Dataset reveal that the annotation workload can be reduced up to 96\% by the proposed approach. Resulting \texttt{uav\_detection\_2} annotations and our codes are publicly available at \href{https://github.com/aybora/Semi-Automatic-Video-Annotation-OGAM}{github.com/aybora/Semi-Automatic-Video-Annotation-OGAM.}
\end{abstract}

\section{Introduction}
\label{sec:intro}

Object detection and tracking algorithms are advanced rapidly during the last decade, especially after realizing the efficiency of Convolutional Neural Networks (CNN) on feature extraction. However, these detection algorithms \cite{Girshick1, Girshick2, yolo, retinanet, yolov3} need a significant amount of annotated data for training which is one of the most important challenges in supervised learning. Similar to object detectors, learning-based trackers \cite{nam2016learning, bertinetto2016fully, danelljan2017eco} also need annotated data. Besides the learning-based methods, we also need annotated data to evaluate the performance of tracking and detection algorithms. For all these applications, researchers need human annotators to specify the positions of the objects in each frame. Object location is defined by a bounding box; hence, each annotation requires two mouse clicks, onto the top-left and bottom-right of the object. For some large public datasets, it is preferred to use crowd-sourcing methods \cite{su2012} to overcome this enormous effort; however, crowd-sourcing is also a costly method. Moreover, for problem-specific and/or confidential datasets, like drone detection and tracking on IR videos, crowd-sourcing might not be applicable. For such cases, the researchers would need to annotate their data frame by frame, which requires lots of labor. As a result, automatic and semi-automatic annotation methods are considered as the only solution to this problem and receive significant attention.

For video object detection and tracking scenarios, object trackers might help to reduce the annotation workload. For example, a popular annotation tool EVA \cite{eva} employs KCF tracker to extrapolate initial boundaing boxes in time. However, template matching-based trackers suffer from drift problem, which might lead to erroneous annotations. At the last year's workshop, a study revealed that the published ground-truth annotations have significant errors which introduces an upper bound of 86.4\% for tracking accuracy \cite{koksal2020effect}.   

Besides using trackers, the research efforts are focused on two different approaches to decrease the annotation workload. The first approach is based on weakly supervised learning. In this approach, a researcher only gives the image and object class in that image as an input and the network has to find the corresponding bounding boxes \cite{bilen2014weakly, bilen2015weakly}. On the other hand, the latter group of techniques exploits active learning. In this approach, the computer actively asks humans to annotate some selected bounding boxes during training \cite{yao2012, vijayanarasimhan2011}. Although these algorithms lead to a decrease in human work, they still require a substantial amount of time to annotate all frames. Recently, some research has focused on incremental learning to overcome this problem \cite{adhikari2020iterative, adhikari2018faster, papadopoulos2016we}. In such methods, at each frame, the human annotator only checks whether the inferred bounding boxes are correct or not to decrease the workload.

\begin{figure}[t]
\centering
   \includegraphics[width=0.85\linewidth]{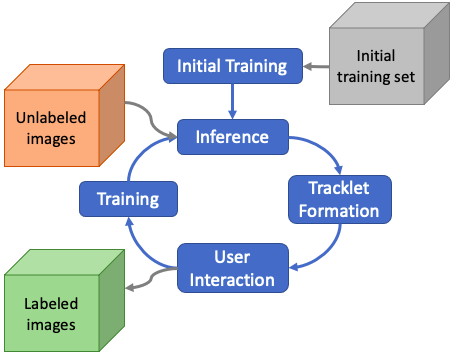}
   \caption{Proposed annotation scheme}
\label{fig:flow}
\end{figure}

In this study, we propose a novel and yet simple method based on incremental learning to annotate the bounding boxes in a video for single object tracking scenario. The general flow of the proposed method is presented in Figure \ref{fig:flow}. We first train an off-the-shelf object detector to perform detection. Then by exploiting the temporal information in a video, we automatically form tracklets, which are consecutive and visually similar detection sets of the same object. Our proposed user interface displays some visual examples for user verification, especially the hard ones, for each tracklet at once. Therefore, the human annotator only needs to check the tracklets instead of each detection, which decreases workload even further. Finally, we enlarge our training set with the detections confirmed by human operator and re-train the object detector for a better performance. 

The proposed method differs from the previous work in following points a) We utilize tracking-by-detection to form tracklets and eliminate false-positives b) We apply a low confidence threshold for detection to increase recall, but a high confidence threshold for track initiation, c) We perform user verification at tracklet-level rather than frame-by-frame, d) We try to annotate the whole set on each iteration rather than annotating batch-by-batch, which helps to converge more rapidly.

\section{Related Work}
\label{sec:rwork}

As more supervised algorithms are employed in object detection, researchers focused more on decreasing the workload of the annotation process. Papadopoulos et al. \cite{papadopoulos2016we} proposed an iterative bounding box annotation method with human verification. For finding initial bounding boxes an improved multiple instance learning method is proposed. Then, an object detector is trained iteratively, and inferred bounding boxes on each frame are evaluated by a human operator. Even though this approach reduces the annotation workload, the method does not exploit temporal information. Therefore an operator have to evaluate every output of the object detector.

Adhikari et al. also worked on iterative bounding box annotation via two studies \cite{adhikari2020iterative, adhikari2018faster}. In this approach an off-the-shelf object detector is trained with a small training set which is annotated manually for initial training. Then, the rest of the study follows the iterative approach \cite{papadopoulos2016we}, i.e., remainder of the training set is inferred with these detectors, verified by human operators and they are used for next training iteration along with previously annotated data. The set is processed batch by batch and at each iteration a batch is completely annotated. Despite Adhikari et al. \cite{adhikari2020iterative} focuses on video annotation, they did not leverage temporal information.

Besides iterative learning approaches, there are also active learning methods that directly need human labeling only for specific frames which are selected automatically depending on the learning performance \cite{konyushkova2018learning, russakovsky2015best, kuznetsova2021efficient}. Even if these algorithms decrease the workload significantly, the operators still have to annotate lots of bounding boxes. 

Although the methods mentioned above are quite efficient in reducing annotation workload, the workload can be decreased further for videos by exploiting temporal data. The simplest way to exploit temporal data is to apply tracking like the well-known annotation tool EVA \cite{eva}; however, tracks might drift and this would lead to erroneous annotations as shown in \cite{koksal2020effect}. 

There are not many fully automatic data annotation methods exploiting temporal data. However, hard example mining techniques in metric learning literature are quite relevant. Jin et al. \cite{jin2018unsupervised} proposed a hard example mining method by exploiting temporal data. The algorithm matches consecutive detections and classifies consistent detections as \textit{pseudo positive}, while non-consistent ones as \textit{hard negative}. The frames on which detector fails to detect the object but tracker (template matching) succeeds are classified as \textit{hard positive}. The training is performed iteratively by weighting all the hard examples. Although this approach is proposed for hard example mining problem, idea of finding consistent detections via matching consecutive detections can be employed for bounding box annotation as well. 

\begin{figure*}[ht]
\centering
   \includegraphics[width=1\textwidth]{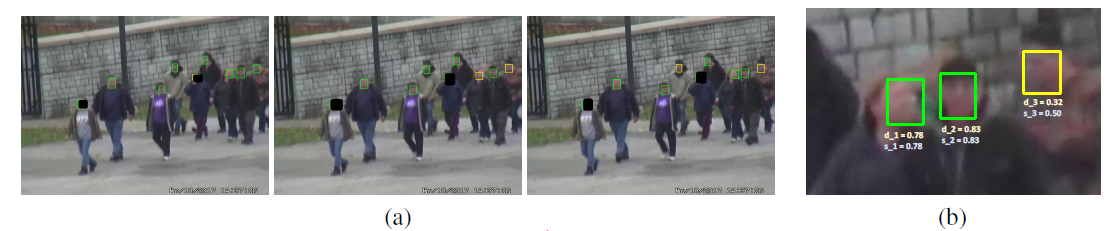}
   \caption{Simple example for the idea of RoyChowdhury et al. \cite{roychowdhury2019automatic}. (a) Pseudo-labels can be seen in either green boxes which represent labels from detector or yellow boxes which are from tracker. (b) $d_i$: detection confidence scores, $s_i$: soft scores.}
\label{fig:chowd}
\end{figure*}

RoyChowdhury et al. \cite{roychowdhury2019automatic} suggested an object detector and tracker combined framework for bounding box annotation, which is also employed to select the hard examples. A baseline object detector, Faster R-CNN \cite{Ren}, is used to generate the pseudo-labels. Then using temporal data with a tracker, MD-Net \cite{nam2016learning}, pseudo-labels are refined by forcing them be temporally consistent. For the training, weighing the labels inferred from detector (soft-examples) or coming from tracker (hard-examples) also improves the performance. The idea of this approach can be seen in Figure \ref{fig:chowd}. Although this approach is the most related one to the proposed work in terms of leveraging temporal data for bounding box annotation, incremental learning might not be possible in this approach, as there is no operator control to eliminate false alarms, which may cause the neural networks to diverge during training.

\section{Proposed Method}
\label{sec:method}

The proposed method is an iterative semi-supervised approach 
for bounding box annotation in a video for single object tracking scenario. At the first step, the selected object detector is trained with a small but fully annotated set. The training step is followed by the inference step as shown in Figure \ref{fig:flow}, and the detection results are merged to form tracklets. Then, each tracklet is evaluated by an operator, and confirmed tracklets are added to the training set to complete the iteration. The next iteration takes place with this enlarged training set. The procedure is simply selecting the true-positive alarms (detection outputs) in a semi-automatic way by enforcing temporal consistency and getting some human supervision. The flowchart of the proposed method is presented in Figure \ref{fig:flow} and each step is further explained in the following subsections.

\noindent\textbf{Tracklet Formation}: The proposed method employs a tracking-by-detection approach to merge the generated alarms to form tracklets. Multiple hypothesis tracking (MHT) \cite{reid1979algorithm} is a popular powerful tool for such problems. The original MHT \cite{reid1979algorithm} utilizes only a trajectory model, while a visual model can be incorporated to reduce ambiguity \cite{kim2015multiple}. We also utilize multiple hypotheses to form tracklets. We employ a constant velocity Kalman Filter as the trajectory model ($T$) as proposed in \cite{kim2015multiple}. We define the visual model ($V$) as the raw pixel measurements, and update the visual model completely on each measurement update. As opposed to the trajectory model, the visual model is much less ambiguous; therefore, each hypothesis ($h$) is allowed to take a single measurement ($d$), which has the highest correlation score, $C(d,h)$, with the visual model (template) if $C(d,h)$ is higher than the selected threshold or the selected measurement lies in the Kalman Filter's gate. Each hypothesis is also allowed to continue with no observation update only if the matching alarm is out of the gate of the Kalman Filter or the correlation score is low. Allowed measurements for different cases are summarized in the Table \ref{tab:hypotheses} as the primary and alternative ones respectively. For the cases in which more than one measurement is allowed, the original hypothesis continues with the primary measurement and a new hypothesis is generated for the alternative measurement.

\begin{table}[ht]
\centering
\caption{Allowed measurements for different cases} 
\label{tab:hypotheses}
\medskip
\begin{tabular}{|l|c|c|c|}
\hline
 & $C<0.5$ & $0.5 \leq C<0.8$ & $C \geq 0.8$ \\
\hline
in gate & $\varnothing$, $d$ & $d$ & $d$ \\ 
\hline
not in gate & $\varnothing$ & $\varnothing$, $d$ & $d$ \\ 
\hline
\end{tabular}
\end{table}

Low and moderate visual similarity scores ($C<0.8$) indicate either fail to detect the object or appearance changes. In case of failure of detection and low visual similarity ($C<0.5$) we need a small gate to avoid from generating an unnecessary hypothesis. In case of appearance changes and low visual similarity ($C<0.5$), gate must be large enough to cover the detection and prevent tracklet to be broken. In moderate visual similarity and failure of detection case, we require a small gate to avoid mismatches. In moderate visual similarity and appearance change case, a small gate help to avoid to generate an unnecessary hypothesis. In short, uncertainty in the position should be reduced as much as possible while keeping the true detections in the gate. To make the constant velocity model fit better to measurement dynamics so to reduce uncertainty, we take camera movement into account by using a Kalman Filter with an input vector:
\begin{equation}
    \begin{gathered}
        x_{k+1} = F * x_k + B * u_k + w_k\\
        y_k = H * x_k + v_k    
    \end{gathered}
    \label{eq:1}
\end{equation}
where state vector $x_k$ is defined as the position and velocity of the object, $u_k$ is the displacement between frames $k$ and $k+1$, $w_k$ is the process noise, $y_k$ is the observation, and $v_k$ is the measurement noise. We use two separate Kalman Filters for horizontal and vertical axes. For both axes we use following state transition matrix ($F$), input matrix ($B$), measurement matrix ($H$), process noise covariance ($\Theta$) and measurement noise covariance ($R$):
\begin{equation}
    \begin{gathered}
        F = \begin{bmatrix} 1 & t \\ 0 & 1 \end{bmatrix}, 
        B = \begin{bmatrix} 1 \\ 0 \end{bmatrix}, 
        \Theta = \begin{bmatrix} t^3/3 & t^2/2 \\ t^2/2 & t \end{bmatrix} * \sigma^2_w
        \\
        H = \begin{bmatrix} 1 & 0 \end{bmatrix},
        R = \sigma^2_v
    \end{gathered}
    \label{eq:2}
\end{equation}
where $t$ is the sampling time between two frames. We estimate velocity per frame by using $t=1$, and we use $\sigma^2_w = 0.005$ and $\sigma^2_v = 2.25$.

We assume that the motion between two frames is translational which corresponds to pan and tilt of the camera. To find the camera movement we apply phase correlation.

Besides the trajectory model, characteristics of the detector is also very important for reducing annotation workload. Fail to detect the object would result in break of the tracklet. To avoid false-negatives, we apply a quite low confidence threshold  ($\theta_L$) on detector's objectness score ($P_d$) for generating alarms. However, to reduce the number of hypotheses, we initiate a new hypothesis only for the detections with high confidence scores ($\theta_H$). 

To keep the number of hypotheses under control we also need a strategy to delete and merge the hypotheses. We indicate the significance of a hypothesis with its average objectness score over last few frames. The hypotheses, whose average objectness score ($P_{avg}$) is below a certain threshold, ($\theta_{avg}$) are deleted. When a tracklet is updated with no measurement, the objectness score at that frame is set to zero. Setting objectness score for no measurement update to zero helps us to delete hypotheses which do not get measurement update for a few frames as their average objectness score decrease on each no measurement update. 

Since the visual model determines the measurement that will be used for the update, two different hypotheses with the same visual model  will be updated with the same measurements and will converge to identical tracklets at the end. Therefore, whenever two different hypotheses take the same measurement, the hypothesis with the highest average objectness score is kept and the rest is deleted. The algorithmic flow of the tracklet formation approach is presented in Algorithm \ref{alg:tracklet}. Note that the performance of the employed tracklet formation method mainly affects the workload of human operator, not the annotation accuracy.

\begin{figure}[ht]
\centering
   \includegraphics[width=1\linewidth]{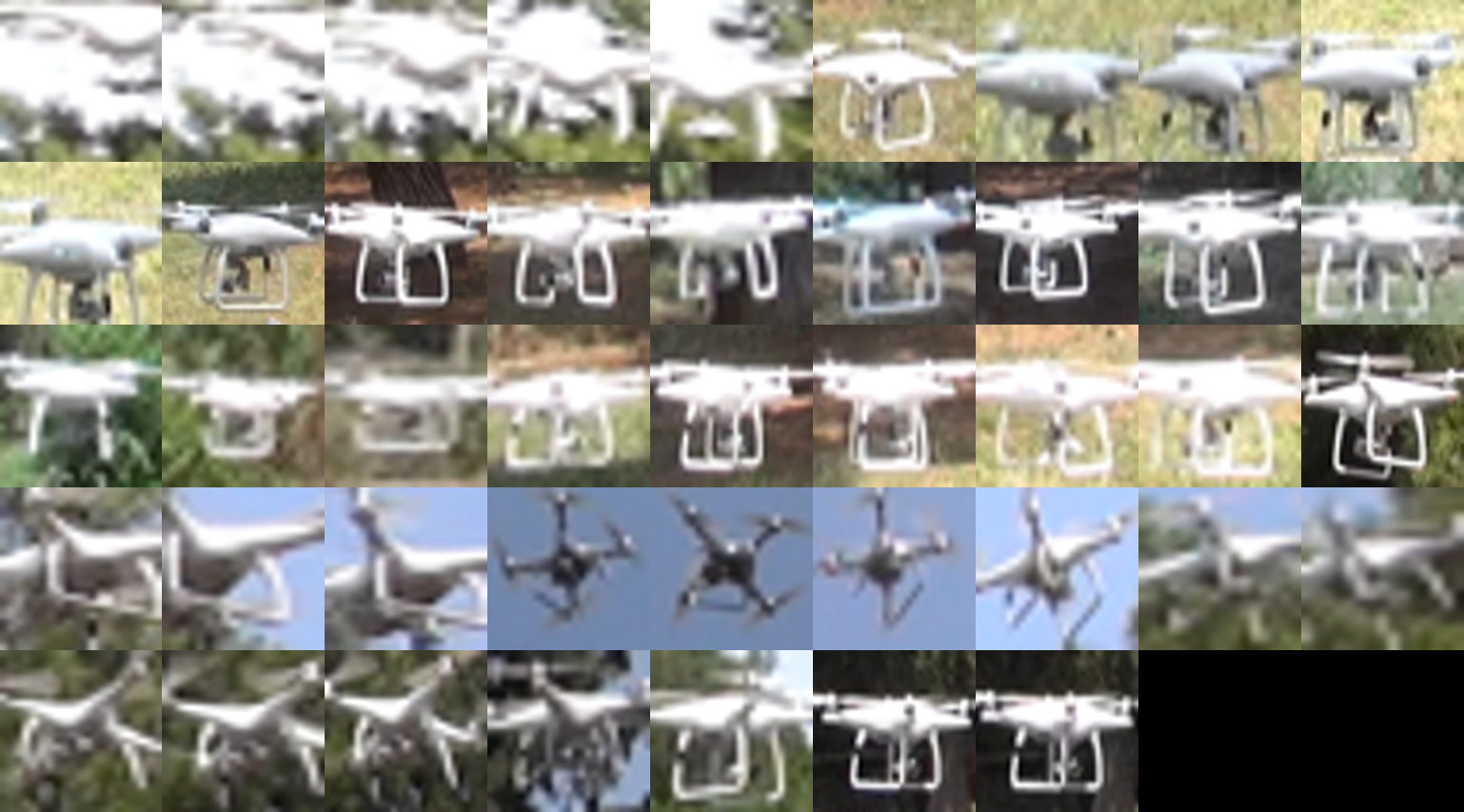}
   \caption{Sample tracklet evaluation screen for $N = 7$}
\label{fig:interface}
\end{figure}

\begin{algorithm}[ht]
\caption{Tracklet Formation}
\label{alg:tracklet}
\begin{algorithmic}
\STATE 1. Select the alarms $d$ with $ P_d > \theta_{L}$
\STATE 2. Perform measurement update:
\FOR {each hypothesis $h$}
    \STATE Select the matching alarm,  $d^* = \argmax_d C(d,h)$
    \STATE Update $V_h$ and $T_h$ with the primary measurement
    \STATE Generate a new hypothesis with alternative measurement, if any
\ENDFOR
\STATE 3. Generate a new hypothesis for each alarm with $ P_d > \theta_{H}$ and not assigned to a hypothesis
\STATE 4. Merge the hypotheses matched with the same alarm
\STATE 5. Remove hypotheses with $ P_{avg}< \theta_{avg}$
\end{algorithmic}
\end{algorithm}
\noindent\textbf{User Interaction}: Once the tracklets are formed, a human operator is asked to evaluate tracklets. A sample tracklet evaluation screen is shown in Fig. \ref{fig:interface}. For  each  tracklet, N samples which have equal temporal spacing among the tracklet are selected. In between those samples, instances having (1) lowest objectness score, (2) lowest correlation score, (3) lowest average objectness score, (4) highest distance to estimated position, and (5) highest temporal distance to other displayed instances are presented to the operator for evaluation. The operator is asked to accept or reject each sample in temporal order. Once a sample is accepted, other samples up to the next rejection are accepted and vice versa. With this approach, for the best case, each tracklet can be evaluated with two clicks (accept or reject) on the first and last samples.

\noindent\textbf{Incremental Learning}: As the iterations take place, we add only the frames which contains a measurement from a user confirmed tracklets to the set of labeled frames and train the detector with this enlarged set. When combined with the proposed user interaction strategy, this approach introduces a limitation for annotating frames containing multiple objects. As user interaction step is not designed to handle missing detections and any frame containing a confirmed object is added to the training set, if there are multiple objects in a frame, objects which are failed to detect will not be annotated. However, as we are focused on bounding box annotation for single object tracking scenario, we leave the solution of this limitation as a future work.

\noindent\textbf{Incremental Annotation}: As the iterations take place, there is no need to re-evaluate the tracklets overlapping with previously accepted/rejected ones. Overlapping instances of those tracklets with the previously accepted/rejected ones are ignored during operator evaluation. As a special case, for single object detection (or tracking) if a frame is already annotated, there is no need to re-prompt that frame for operator evaluation. Since the experiments are performed on AUTH \texttt{uav\_detection} subset \cite{mademlis1,mademlis2}, which is a single object detection case, on each iteration only the frames without annotation are evaluated.
\section{Experiments}
\label{sec:experiments}


Throughout the experiments, YOLO-v3 \cite{yolov3} is used as the baseline detector. To demonstrate the effectiveness of the proposed method, we re-annotated the \texttt{uav\_detection} subset and annotated the \texttt{uav\_detection\_2} subset of AUTH Multidrone set \cite{mademlis1,mademlis2} which is available at \footnote{\href{http://poseidon.csd.auth.gr/LAB_PROJECTS/MULTIDRONE/AUTH_MULTIDRONE_Dataset.html}{poseidon.csd.auth.gr/LAB\_PROJECTS/MULTIDRONE/AUTH\_MUL\\TIDRONE\_Dataset.html}}. We also compared operator workload against alternative methods. 

The results are are reported in terms of recall percent (Rec), number of False Alarms Before Click (FA-B), number of False Alarms After Click (FA-A), number of Clicks, number of Annotated Frames (Ann \#), percentage of Annotated Frames over All Frames with a Drone (Ann \%). (Rec, FA-B, FA-A are presented if the ground-truth is available).

\textbf{Enlarging the training set} is a quite common need for supervised methods. As the learning-based systems are employed more, more data is obtained every day which can be utilized for training to enhance the performance. On the other hand, for a fixed set one might prefer to annotate a part of the available data and annotate the rest with the help of the trained detector. To simulate these two cases, we have selected two videos (videos DSC\_6303 and DSC\_6304) having a total number of 4803 frames from AUTH \texttt{uav\_detection} subset as the initial training set, and re-annotated the rest of the subset with the proposed method. We simulated user clicks using the original annotations such that if the intersection over union of the proposed and the original annotation is larger than $\theta_{IoU}$ ($\dagger$) the user is assumed click accept and reject otherwise. As shown in Table \ref{tab:iterwith5000}, in 7 iterations, we were able to annotate 35960 frames with only 2712 clicks. As each bounding box is defined by two clicks, the annotation effort for the proposed method corresponds to 3.7\% of unaided case. To annotate the whole set with this approach, we require 4803 initial annotations, 2712 tracklet evaluation clicks, and 1005 manual annotations for the remaining boxes, which corresponds to 82.84\% workload reduction.


\textbf{Initiating training set} might also be performed by annotating some sample frames from each video manually rather than completely annotating some videos manually and leaving the rest. To simulate that case, we initiated training set by selecting the original annotations of 1 frame for every 100 frames, by uniform temporal sampling. Then, we re-annotate rest of the training set (99\%) with the proposed method. As shown in Table \ref{tab:iterwith100}, in 5 iterations we were able to annotate 40469 frames with only 1518 clicks which corresponds to an annotation effort of 1.9\% of unaided case. Accounting initial annotations (421 boxes) and manual annotation of remaining frames after iterations, workload reduction is 94.14\% for this case. When compared to the first experiment which has 4803 annotations initially, selecting frames for initial annotation from a temporally uniform distribution, helps to converge faster with much less effort.

\begin{figure}[ht!]
\centering
   \includegraphics[width=1\linewidth]{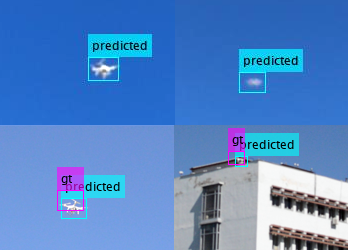}
   \caption{At the top, examples of inferred boxes (predicted) which are tabulated as false alarms in Table \ref{tab:iterwith5000} and \ref{tab:iterwith100}; at the bottom examples for the annotations which should be evaluated with $\theta_{IoU} = 0.2$ instead of $\theta_{IoU} = 0.5$ due to the shown incorrect original ground-truth (gt). Displayed frames are (a) 8445 of 00001, (b) 7912 of 00002, (c) 3682 of DSC\_6299, (d) 1222 of DSC\_6304.}
\label{fig:falarms}
\end{figure}

FA-A columns of Tables \ref{tab:iterwith5000} and \ref{tab:iterwith100} are not false alarms for real; they are either incorrect ground-truth, or some blurry frames which the operator decided not to annotate.
Typical visual examples are given in Figure \ref{fig:falarms}. Such errors are known as missing and shifted box errors which reduce the performance of the detector and would mislead the performance measurement \cite{koksal2020effect}.
\newline
\begin{table}[ht]
\centering
\caption{Results of enlarging training set experiment} 
\label{tab:iterwith5000}
\medskip
\begin{tabular}{|l|c|c|c|c|c|c|}
\hline
Iter & Rec & FA-B & FA-A & Click & Ann \# & Ann \% \\
\hline
1 & 66.81 & 1809 & 6 & 757 & 26162 & 70.77 \\ 
2 & 82.11 & 573 & 6 & 737 & 31694 & 85.74 \\ 
3 & 85.93 & 437 & 6 & 533 & 33074 & 89.47 \\ 
4 & 88.74 & 139 & 6 & 410 & 34136 & 92.34 \\ 
5 & 96.23 & 56 & 6 & 99 & 35666 & 96.48 \\ 
6 & 96.92 & 53 & 6 & 90 & 35899 & 97.11 \\ 
7 & 97.09 & 103 & 6 & 86 & 35960 & 97.28 \\ 
\hline
\multicolumn{3}{|c|}{Total Clicks: 2712} & \multicolumn{4}{|c|}{Annotated: 35960 (97.28 \%)} \\
\hline
\end{tabular}

\end{table}

\begin{table}[ht]
\centering
\caption{Results of initiating training set experiment} 
\label{tab:iterwith100}
\medskip
\begin{tabular}{|l|c|c|c|c|c|c|}
\hline
Iter & Rec & FA-B & FA-A & Click & Ann \# & Ann \% \\
\hline
1 & 88.76 & 1795 & 47 & 671 & 37823 & 90.58 \\ 
2 & 92.40 & 798 & 47 & 509 & 39345 & 94.22 \\ 
3 & 96.03 & 178 & 47 & 153 & 40235 & 96.35 \\ 
4 & 96.47 & 114 & 47 & 93 & 40412 & 96.78 \\ 
5 & 96.61 & 277 & 47 & 92 & 40469 & 96.91 \\ 
\hline
\multicolumn{3}{|c|}{Total Clicks: 1518} & \multicolumn{4}{|c|}{Annotated: 40469 (96.91 \%)} \\
\hline
\end{tabular}

\end{table}

($\dagger$) It should be noted that the last 3 rows of Tables \ref{tab:iterwith5000} and \ref{tab:iterwith100} are evaluated with $\theta_{IoU} = 0.2$, while the rest with $\theta_{IoU} = 0.5$ for simulating user action, as some original annotations are slightly shifted (shifted box error).

\noindent\textbf{Annotating AUTH UAV\_Detection\_2 Subset:}
Through the first two experiments, we have noticed that the original annotations might be incorrect, especially might contain box shift errors. As the original annotations are slightly shifted, the proposed annotations have low IoU with them, which results in more clicks than needed. To demonstrate the effectiveness of the proposed method in a more fair case, we also annotated AUTH \texttt{uav\_detection\_2} subset. For this experiment, we use \texttt{uav\_detection} subset for initial training. As shown in Table \ref{tab:uav2}, in 4 iterations we were able to annotate 22262 frames with total number of 810 clicks, which corresponds to 96.25\% workload reduction with respect to unaided case. This performance enhancement might be due to box shift error in original annotations as mentioned above, but it might be due to the larger initial training set as well. However, as seen in Table \ref{tab:uav2} at the first iteration we are able to annotate only 43\% of the frames, while in Tables \ref{tab:iterwith5000} and \ref{tab:iterwith100} at the first iteration semi-automatic annotation rates are 70\% and 90\% respectively, supporting the performance enhancement due to getting rid of box shift errors.

\begin{table}[ht]
\centering
\caption{Annotation Effort for \textbf{UAV\_Detection\_2 Subset} } 
\label{tab:uav2}
\medskip
\begin{tabular}{|l|c|c|c|}
\hline
Iter & Click & Ann \# & Ann \% \\
\hline
1 & 280 & 9765 & 43.00 \\ 
2 & 231 & 18771 & 82.66 \\ 
3 & 195 & 21828 & 96.12 \\ 
4 & 104 & 22262 & 98.03 \\ 
\hline
\multicolumn{2}{|c|}{Total Clicks: 810} & \multicolumn{2}{|c|}{Ann (\%): 98.03} \\
\hline
\end{tabular}
\end{table}

\noindent\textbf{Comparison with Alternative Methods:}
RoyChowdhury et al. \cite{roychowdhury2019automatic} increased their face detection performance from a baseline of 15.66 AP to 20.65 on WIDER-Face unlabeled CS6 dataset \cite{yang2016wider} by automatic annotation. For pedestrian detection, they use a part of BDD-100K 
dataset \cite{yu2018bdd100k} 
as initial training set and automatically annotated the rest wit a fully automatic method which increased their AP from 15.21 to 28.43 in automatically annotated set. We increased AP for drone detection from 69.73 to 80.73 for enlarging training set scenario.

Workload reduction can be calculated as the decrease on a number of the clicks needed to annotate the dataset completely. Comparison with different approaches on literature is presented in Table \ref{tab:workload}. As seen in the Table, the proposed approach is able to reduce workload up to 15\% more by exploiting temporal data and tracklet level user interaction.

\begin{table}[ht]
\centering
\caption{Comparison of proposed annotation method with alternative approaches in terms of workload reduction}
\label{tab:workload}
\medskip
\begin{tabular}{|l|c|}
\hline
Method & Reduction (\%) \\
\hline
Adhikari et al. \cite{adhikari2020iterative} (Best) & 80.56 \\
Adhikari et al. \cite{adhikari2018faster} (Best) & 81.26 \\
Enlarging Training Set (Ours) & 82.84 \\
Initiating Training Set (Ours) & 94.14 \\
Annotating New Set (Ours) & 96.25 \\
\hline
\end{tabular}
\end{table}
\section{Conclusions}
\label{sec:conclusion}

The proposed semi-automatic method for bounding box annotation for single object tracking scenario is shown to reduce the human workload by 82-96\% for two different enlarging training set scenarios. For initiating training set scenario, if some videos are fully annotated and then the proposed method is applied, workload reduction is 82\%. However, if the initial labeling effort is spent on temporally uniformly sampled frames, workload reduction increases up to 94\%. It is obvious that selecting more temporally spaced frames for initial manual annotation helps to increase the variation of the initial set and having an initial set with more variation helps to converge faster with less workload. The proposed method reduces the workload more than the other semi-automatic methods in literature.

The proposed method has some limitations. First of all, even if the incremental learning approach is also applicable for bounding box annotation on independent still images as demonstrated in \cite{adhikari2018faster}, the proposed method is applicable only for videos. Moreover, human interaction and incremental learning approaches utilized by the proposed method might fail for the videos containing multiple objects. For such scenarios, a more suitable human interaction step should be designed. 

\section{Acknowledgements}

\ificcvfinal
This study is partially funded by ASELSAN Inc.

\else Acknowledgements will be here in the final version.

\fi

{\small
\bibliographystyle{ieee_fullname}
\bibliography{refs}
}

\end{document}